\documentclass[10pt,twocolumn,letterpaper]{article}

\usepackage{cvpr}
\usepackage{times}
\usepackage{cuted}
\usepackage{epsfig}
\usepackage{graphicx}
\usepackage{amsmath}
\usepackage{amssymb}
\usepackage{subfigure}
\usepackage{dsfont}
\usepackage{array,multirow}
\usepackage{float}

\usepackage{soul,color}

\newcommand{\norm}[1]{\left\lVert#1\right\rVert}
\usepackage{graphicx,calc}
\newlength\myheight
\newlength\mydepth
\settototalheight\myheight{Xygp}
\newcommand*\inlinegraphics[1]{%
  \settototalheight\myheight{Xygp}%
  \settodepth\mydepth{Xygp}%
  \raisebox{-\mydepth}{\includegraphics[height=\myheight]{#1}}%
}
\usepackage[pagebackref=true,breaklinks=true,letterpaper=true,colorlinks,bookmarks=false]{hyperref}

\def\cvprPaperID{7874} 

\cvprfinalcopy
\ifcvprfinal\fi
\begin{document}

\title{TITAN: Future Forecast using Action Priors}
\author{Srikanth Malla~~~~~~~~~~~~~~~~~~Behzad Dariush~~~~~~~~~~~~~~~~~~Chiho Choi\\
Honda Research Institute USA\\
{\tt\small \{smalla, bdariush, cchoi\}@honda-ri.com }
}

\maketitle
\begin{abstract}
We consider the problem of predicting the future trajectory of scene agents from egocentric views obtained from a moving platform. This problem is important in a variety of domains, particularly for autonomous systems making reactive or strategic decisions in navigation. In an attempt to address this problem, we introduce TITAN (Trajectory Inference using Targeted Action priors Network), a new model that incorporates prior positions, actions, and context to forecast future trajectory of agents and future ego-motion.  In the absence of an appropriate dataset for this task, we created the TITAN dataset that consists of 700 labeled video-clips (with odometry) captured from a moving vehicle on highly interactive urban traffic scenes in Tokyo.  Our dataset includes 50 labels including vehicle states and actions, pedestrian age groups, and targeted pedestrian action attributes that are organized hierarchically corresponding to atomic, simple/complex-contextual, transportive, and communicative actions.  To evaluate our model, we conducted extensive experiments on the TITAN dataset, revealing significant performance improvement against baselines and state-of-the-art algorithms.  We also report promising results from our Agent Importance Mechanism (AIM), a module which provides insight into assessment of perceived risk by calculating the relative influence of each agent on the future ego-trajectory. The dataset is available at \url{https://usa.honda-ri.com/titan}
\end{abstract}

\section{Introduction}
The ability to forecast future trajectory of agents (individuals, vehicles, cyclists, etc.) is paramount in developing navigation strategies in a range of applications including motion planning and decision making for autonomous and cooperative (shared autonomy) systems.
\begin{figure}[t!]
    \centering
    \includegraphics[width=0.45\textwidth]{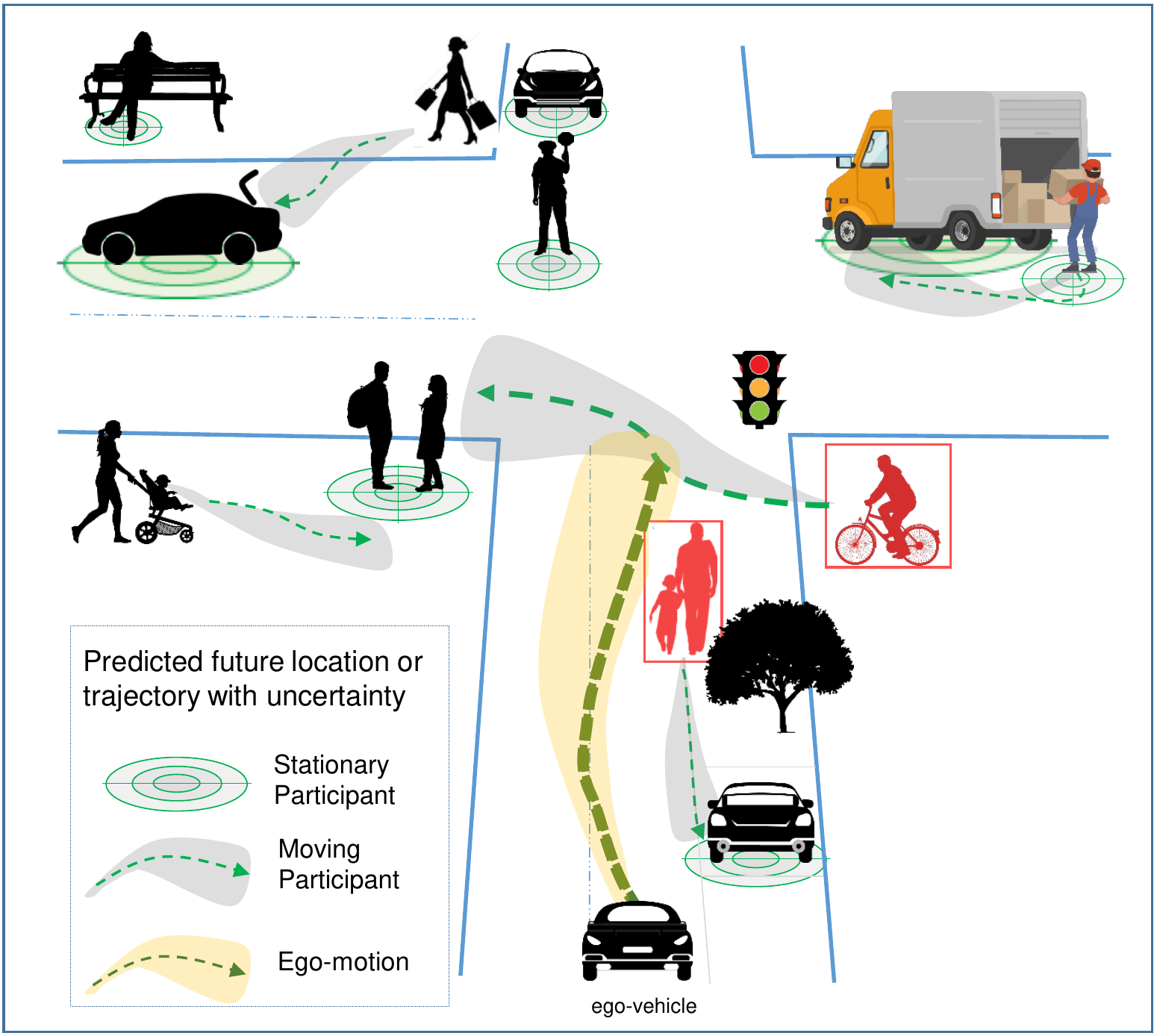}
    \caption{Our goal is to predict the future trajectory of agents from egocentric views obtained from a moving platform.  We hypothesize that prior actions (and implicit intentions) play an important role in future trajectory forecast.  To this end, we develop a model that incorporates prior positions, actions, and context to forecast future trajectory of agents and future ego-motion.  This figure is a conceptual illustration that typifies navigation of ego-vehicle in an urban scene, and how prior actions/intentions and context play an important role in future trajectory forecast. We seek to also identify agents (depicted by the red bounding box) that influence future ego-motion through an Agent Importance Mechanism (AIM) . 
   }
    \label{fig:block_diagram}
    \vspace{-0.51cm}
\end{figure}
We know from observation that the human visual system possesses an uncanny ability to forecast behavior using various cues such as experience, context, relations, and social norms. For example, when immersed in a crowded driving scene, we are able to reasonably estimate the intent, future actions, and future location of the traffic participants in the next few seconds. This is undoubtedly attributed to years of prior experience and observations of interactions among humans and other participants in the scene. To reach such human level ability to forecast behavior is part of the quest for visual intelligence and the holy grail of autonomous navigation, requiring new algorithms, models, and datasets.

In the domain of behavior prediction, this paper considers the problem of future trajectory forecast from egocentric views obtained from a mobile platform such as a vehicle in a road scene. This problem is important for autonomous agents to assess risks or to plan ahead when making reactive or strategic decisions in navigation. Several recently reported models that predict trajectories incorporate social norms, semantics, scene context, etc. The majority of these algorithm are developed from a stationary camera view in surveillance applications, or overhead views from a drone.

The specific objective of this paper is to develop a model that incorporates prior positions, actions, and context to simultaneously forecast future trajectory of agents and future ego-motion.  
In a related problem, the ability to predict future actions based on current observations has been previously studied in~\cite{lan2014hierarchical,soomro2016predicting,singh2017online,singh2018predicting,sun2019relational}. 
However, to the best of our knowledge, action priors have not been used in forecasting future trajectory, partly due to  a lack of an appropriate dataset.   A solution to this problem can help address the challenging and intricate scenarios that capture the interplay of observable actions and their role in future trajectory forecast.  For example, when the egocentric view of a mobile agent in a road scene captures a delivery truck worker closing the tailgate of the truck, it is highly probable that the worker's future behavior will be to walk toward the driver side door. Our aim is to develop a model that uses such action priors to forecast trajectory.

The algorithmic contributions of this paper are as follows.   We introduce TITAN (Trajectory Inference using Targeted Action priors Network), a new model that incorporates prior positions, actions, and context to simultaneously forecast future trajectory of agents and future ego-motion.  Our framework introduces a new interaction module to handle dynamic number of objects in the scene. While modeling pair-wise interactive behavior from all agents, the proposed interaction module incorporates actions of individuals in addition to their locations, which helps the system to understand the contextual meaning of motion behavior. In addition, we propose to use multi-task loss with aleatoric homoscedastic uncertainty~\cite{kendall2018multi} to improve the performance of multi-label action recognition. For ego-future, Agent Importance Mechanism (AIM) is presented to identify objects that are more relevant for ego-motion prediction.

Apart from algorithmic contributions, we introduce a novel dataset, referred to as TITAN dataset, that consists of 700 video clips captured from a moving vehicle on highly interactive urban traffic scenes in Tokyo.  The pedestrians in each clip were labeled with various action attributes that are organized hierarchically corresponding to atomic, simple/complex contextual, transportive, and communicative actions.   The action attributes were selected based on commonly observed actions in driving scenes, or those which are important for inferring intent (\textit{e.g.}, waiting to cross).  We also labeled other participant categories, including vehicle category (4 wheel, 2 wheel), age-groups, and vehicle state.   The dataset contains synchronized  ego-motion information from an IMU sensor.  To our knowledge, this is the only comprehensive and large scale dataset suitable for studying action priors for forecasting the future trajectory of agents from ego-centric views obtained from a moving platform. Furthermore, we believe our dataset will contribute to advancing research for action recognition in driving scenes.

\begin{figure*}[t]
    \centering
    \includegraphics[width=\textwidth]{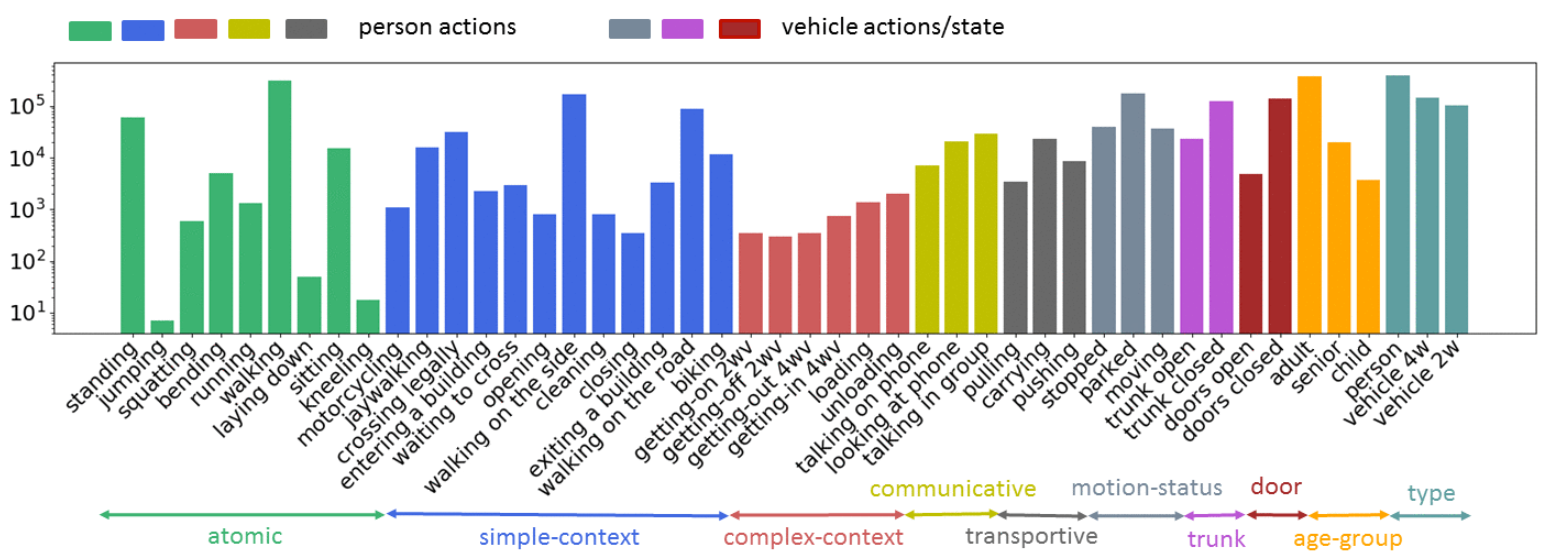}
    \caption{Distribution of labels sorted according to person actions, vehicle actions/state, and other labels such as age groups and types.}
    \label{fig:titan_dataset}
    \vspace{-0.51cm}
\end{figure*}
\section{Related Work}
\subsection{Future Trajectory Forecast} 
\noindent
\textbf{Human Trajectory Forecast} Encoding interactions between humans based on their motion history has been widely studied in the literature. Focusing on input-output time sequential processing of data, recurrent neural network (RNN)-based architectures have been applied to the future forecast problem in the last few years ~\cite{social_lstm,lee2017desire,social_gan,xu2018encoding,zhang2019sr}. More recently, RNNs are used to formulate a connection between agents with their interactions using graph structures~\cite{social_attention,ma2019trafficpredict}. However, these methods suffer from understanding of environmental context with no or minimal considerations of scene information. To incorporate models of human interaction with the environment, \cite{xue2018ss} takes local to global scale image features into account. More recently, \cite{choi2019looking} visually extracts relational behavior of humans interacting with other agents as well as environments.

\noindent
\textbf{Vehicle Trajectory Forecast} Approaches for vehicle motion prediction have developed following the success of interaction modeling using RNNs. Similar to human trajectory forecast, \cite{deo2018multi,park2018sequence,ma2019trafficpredict,li2019interaction} only consider the past motion history. These methods perform poorly in complex road environments without the guidance of structured layouts.  Although the subsequent approaches~\cite{rhinehart2018r2p2,li2019conditional,choi2019drogon} partially overcome these issues by using 3D LiDAR information as inputs to predict future trajectories, their applicability to current production vehicles is limited due to the higher cost.  Recent methods~\cite{bhattacharyya2018long,hevi_dataset,malla2019nemo} generate trajectories of agents from an egocentric view. 
However, they do not consider interactions between road agents in the scene and the potential influence to the ego-future. In this work, we explicitly model pair-wise interactive behavior from all agents to identify objects that are more relevant for the target agent.

\subsection{Action Recognition}
With the success of 2D convolutions in image classification, frame-level action recognition has been presented in~\cite{karpathy2014large}. Subsequently, \cite{simonyan2014two} separates their framework into two streams: one to encode spatial features from RGB images and the other to encode temporal features from corresponding optical flow. Their work motivated studies that model temporal motion features together with spatial image features from videos. A straightforward extension has been shown in~\cite{tran2015learning,varol2017long}, replacing 2D convolutions by 3D convolutions. To further improve the performance of these models, several research efforts have been provided such as I3D~\cite{i3d} that inflates a 2D convolutional network into 3D to benefit from the use of pre-trained models and 3D ResNet~\cite{r3d} that adds residual connections to build a very deep 3D network. Apart from them, other approaches capture pair-wise relations between actor and contextual features~\cite{acrn_cvpr} or those between pixels in space and in time~\cite{non_local_neuralnets}. More recently, Timeception~\cite{hussein2019timeception} models long range temporal dependencies, particularly focusing on complex actions. 


\subsection{Datasets}
\noindent
\textbf{Future Trajectory}
\label{sec:dataset_futuretraj}
Several influential RGB-based datasets for pedestrian trajectory prediction have been reported in the literature.  These datasets are typically created from a stationary surveillance camera~\cite{ucy_dataset, eth_dataset, actEV_dataset}, or from aerial views obtained from a static drone-mounted camera~\cite{stanford_drone_dataset}. In driving scenes, the 3D point cloud-based datasets~\cite{kitti_dataset, h3d_dataset, lyft_dataset, nuscenes_dataset, waymo_open_dataset, argoverse_dataset} were originally introduced for detection, tracking, etc., but recently used for vehicle trajectory prediction as well. Also, \cite{hevi_dataset, chandra2019traphic} provide RGB images captured from an egocentric view of a moving vehicle and applied to future trajectory forecast problem.  The JAAD~\cite{jaad_dataset}, CMU-UAH~\cite{minguez2018pedestrian}, and PIE~\cite{pie_dataset} datasets are most similar to our TITAN dataset in the sense that they are designed to study the intentions and actions of objects from on-board vehicles. However, their labels are limited to simple actions such as walking, standing, looking, and crossing. These datasets, therefore, do not provide an adequate number of actions to use as priors in order to discover contextual meaning of agents’ motion behavior. To address these limitations, our TITAN dataset provides 50 labels including vehicle states and actions, pedestrian age groups, and targeted pedestrian action attributes that are hierarchically organized as illustrated in the supplementary material.

\noindent
\textbf{Action Recognition}
\label{sec:dataset_action}
A variety of datasets have been introduced for action recognition with a single action label~\cite{hmdb_dataset,ucf_dataset,karpathy2014large,marszalek2009actions,kay2017kinetics} and multiple action labels~\cite{charades,yeung2018every,caba2015activitynet} in videos. Recently released datasets such as AVA~\cite{ava_dataset}, READ~\cite{fontana2018action}, and EPIC-KITCHENS~\cite{Damen2018EPICKITCHENS} contain actions with corresponding localization around a person or object. Our TITAN dataset is similar to AVA in the sense that it provides spatio-temporal localization for each agent with multiple action labels. However, the labels of TITAN are organized hierarchically from primitive atomic actions to complicated contextual activities that are typically observed from on-board vehicles in driving scenes.

\begin{figure*}[t!]
    \centering
    \includegraphics[width=\textwidth]{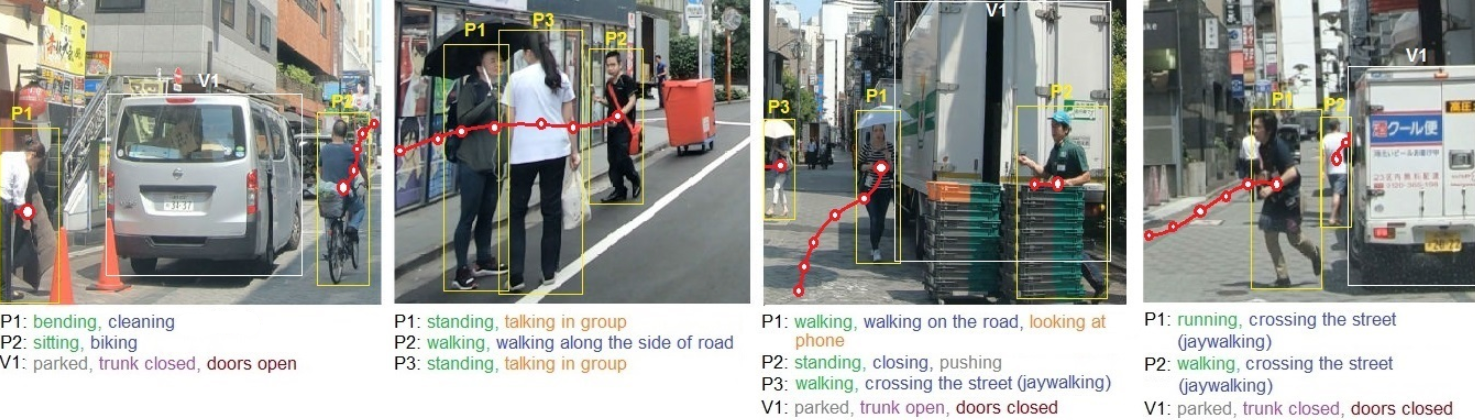}
    \caption{Example scenarios of the TITAN Dataset: a pedestrian bounding box with tracking ID is shown in \protect\inlinegraphics{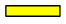}, vehicle bounding box with ID is shown in \protect\inlinegraphics{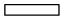}, future locations are displayed in \protect\inlinegraphics{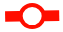}. Action labels are shown in different colors following Figure~\ref{fig:titan_dataset}.}
    \label{fig:titan_scenes}
    \vspace{-0.51cm}
\end{figure*}
\section{TITAN Dataset}
In the absence of an appropriate dataset suitable for our task, we introduce the TITAN dataset for training and evaluation of our models as well as to accelerate research on trajectory forecast.   Our dataset is sourced from 10 hours of video recorded at 60 FPS in central Tokyo.  All videos are captured using a GoPro Hero 7 Camera with embedded IMU sensor which records synchronized odometry data at 100 HZ for ego-motion estimation.  To create the final annotated dataset, we extracted 700 short video clips from the original (raw) recordings.  Each clip is between 10-20 seconds in duration, image size width:1920px, height:1200px and annotated at 10 HZ sampling frequency.  The characteristics of the selected video clips include scenes that exhibit a variety of participant actions and interactions.  

The taxonomy and distribution of all labels in the dataset are depicted in Figure~\ref{fig:titan_dataset}.   The total number of frames annotated is approximately 75,262 with 395,770 persons, 146,840 4-wheeled vehicles and 102,774 2-wheeled vehicles.  This includes 8,592 unique persons and 5,504 unique vehicles.   For our experiments, we use 400 clips for training, 200 clips for validation and 100 clips for testing.   As mentioned in Section~\ref{sec:dataset_futuretraj}, there are many publicly available datasets related to mobility and driving, many of which include ego-centric views.   However, since those datasets do not provide action labels, a meaningful quantitative comparison of the TITAN dataset with respect to existing mobility datasets is not possible. Furthermore, a quantitative comparison with respect to action localization datasets such as AVA is not warranted since AVA does not include ego-centric views captured from a mobile platform.  

In the TITAN dataset, every participant (individuals, vehicles, cyclists, etc.) in each frame is localized using a bounding box. We annotated 3 labels (person, 4-wheeled vehicle, 2-wheeled vehicle), 3 age groups for person (child, adult, senior), 3 motion-status labels for both 2 and 4-wheeled vehicles, and door/trunk status labels for 4-wheeled vehicles.   For action labels, we created 5 mutually exclusive person action sets organized hierarchically (Figure~\ref{fig:titan_dataset}).  In the first action set in the hierarchy, the annotator is instructed to assign exactly one class label among 9 atomic whole body actions/postures that describe primitive action poses such as sitting, standing, standing, bending, etc.  The second action set includes 13 actions that involve single atomic actions with simple scene context such as jaywalking, waiting to cross, etc.  The third action set includes 7 complex contextual actions that involve a sequence of atomic actions with higher contextual understanding, such as getting in/out of a 4-wheel vehicle, loading/unloading, etc.   The fourth action set includes 4 transportive actions that describe the act of manually transporting an object by carrying, pulling or pushing.   Finally, the fifth action set includes 4 communicative actions observed in traffic scenes such as talking on the phone, looking at phone, or talking in groups.   In each action sets 2-5, the annotators were instructed to assign `None' if there is no label.   This hierarchical strategy was designed to produce unique (unambiguous) action labels while reducing the annotators' cognitive work-load and thereby improving annotation quality. The tracking ID's of all localized objects are associated within each video clip. Example scenarios are displayed in Figure~\ref{fig:titan_scenes}.
\begin{figure*}[t]
    \centering
    \includegraphics[width=\textwidth]{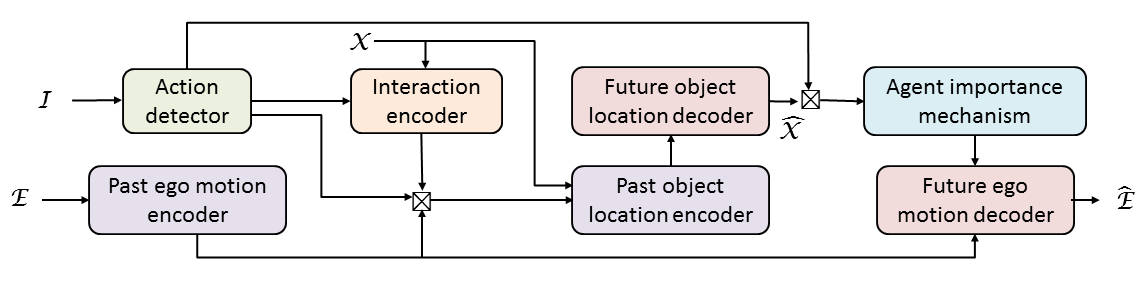}
     \vspace{-0.71cm}
    \caption{The proposed approach predicts the future motion of road agents and ego-vehicle in egocentric view by using actions as a prior. The notation $\mathcal{I}$ represents input images, $\mathcal{X}$ is the input trajectory of other agents, $E$ is the input ego-motion, $\mathcal{\hat{X}}$ is the predicted future trajectory of other agents, and $\hat{E}$ is the predicted future ego-motion.}
    \label{fig:block_diagram}
    \vspace{-0.51cm}
\end{figure*}
\section{Methodology}

Figure~\ref{fig:block_diagram} shows the block diagram of the proposed TITAN framework.A sequence of image patches $\mathcal{I}^i_{t=1:T_{obs}}$ is obtained from the bounding box\footnote{We assume that the bounding box detection using past images is provided by the external module since detection is not the scope of this paper.} $x^i=\{c_u,c_v,l_u,l_v\}$ of  agent $i$ at each past time step from 1 to $T_{obs}$, where $(c_u,c_v)$ and $(l_u,l_v)$ represent the center and the dimension of the bounding box, respectively. The proposed TITAN framework requires three inputs as follows: $\mathcal{I}^i_{t=1:T_{obs}}$ for the action detector, $x^i_t$ for both the interaction encoder and past object location encoder, and $e_t=\{\alpha_t,{\omega}_t\}$ for the ego-motion encoder where $\alpha_t$ and $\omega_t $ correspond to the acceleration and yaw rate of the ego-vehicle at time $t$, respectively. During inference, the multiple modes of future bounding box locations are sampled from a bi-variate Gaussian generated by the noise parameters, and the future ego-motions $\hat{e_t}$ are accordingly predicted, considering the multi-modal nature of the future prediction problem. 

Henceforth, the notation of the feature embedding function using multi-layer perceptron (MLP) is as follows: $\Phi$ is without any activation, and $\Phi_r$,  $\Phi_t$, and $\Phi_s$ are associated with ReLU, tanh, and a sigmoid function, respectively. 
\subsection{Action Recognition}
\label{sec:action_rec}
We use the existing state-of-the-art method as backbone for the action detector. We finetune single-stream I3D~\cite{i3d} and 3D ResNet~\cite{r3d} architecture pre-trained on Kinetics-600~\cite{carreira2018short}. The original head of the architecture is replaced by a set of new heads (8 action sets of TITAN except age group and type) for multi-label action outputs. The action detector takes $\mathcal{I}^i_{t=1:T_{obs}}$ as input, which is cropped around the agent $i$. 
Then, each head outputs an action label including a ‘None’ class if no action is shown. From our experiments, we observed that certain action sets converge faster than others. This is due in part because some tasks are relatively easier to learn, given the shared representations. Instead of tuning the weight of each task by hand, we adopt the multi-task loss in \cite{kendall2018multi} to further boost performance of our action detector. Note that each action set of the TITAN dataset is mutually exclusive, thus we consider the outputs are independent to each other as follows:
\begin{equation}
    p(y_m,..,y_n|f(\mathcal{I}))=\prod_{i=m}^{n} p(y_i|f(\mathcal{I})),
    \label{eq:multi_task_out}
\end{equation}
where $y_i$ is the output label of $i^{th}$ action set and $f$ is the action detection model. Then, multi-task loss is defined as:
\begin{equation}
    \mathcal{L}_a=\sum_{i=m}^{n}\frac{ce(\widehat{cls}_i, {cls}_i)}{\sigma_i^2}+log\ \sigma_i,
    \label{eq:multi_task_loss}
\end{equation}
where $ce$ is the cross entropy loss between predicted actions $\widehat{cls}_i$ and ground truth ${cls}_i$ for each label $i=m:n$. Also, $\sigma_i$ is the task dependent uncertainty (aleatoric homoscedastic). In practice, the supervision is done separately for vehicles and pedestrians as they have different action sets. The efficacy of the multi-task loss is detailed in the supplementary material, and the performance of the action detector with different backbone is compared in Table~\ref{tbl:action_recog}. 

\subsection{Future Object Localization}
\label{sec:fol_methodology}
\begin{figure}[t!]
    \centering
    \includegraphics[width=0.47\textwidth]{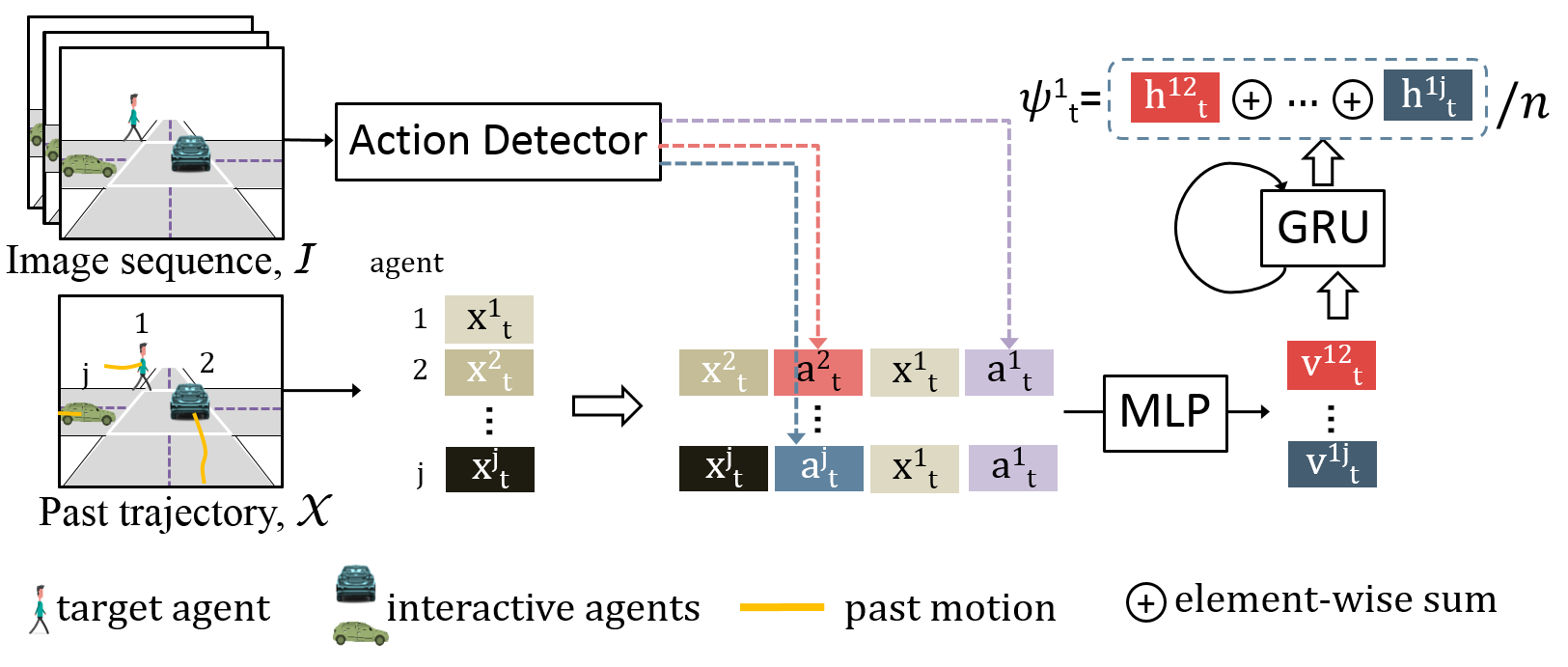}
    \caption{Interaction encoding for agent $i$ against others at time $t$.}
    \label{fig:interaction_encoder}
    \vspace{-0.51cm}
\end{figure}

Unlike existing methods, we model the interactions using the past locations of agents conditioned on their actions, which enables the system to explicitly understand the contextual meaning of motion behavior. At each past time step $t$, the given bounding box $x^i_t=\{c_u,c_v,l_u,l_v\}_t$ is concatenated with the multi-label action vector $a^i_t$. We model the pair-wise interactions between the target agent $i$ and all other agents $j$ through MLP, $v^{ij}_t=\Phi_r(x^{i}_t\boxtimes a^{i}_t\boxtimes x^{j}_t \boxtimes a^{j}_t)$ where $\boxtimes$ is a concatenation operator. The resulting interactions $v_t^{ij}$ are evaluated through the dynamic RNN with GRUs to leave more important information with respect to the target agent, $h^{i(j+1)}_t = GRU(v^{ij}_t, h_t^{ij};W_{\textnormal{INT}})$, where $W_{\textnormal{INT}}$ are the weight parameters. Note that we pass the messages of instant interaction with each agent at time $t$, which enables us to find their potential influence at that moment. Then, we aggregate the hidden states to generate interaction features $\psi_{t}^i=\frac{1}{n}\sum_i h^{ij}_t$ for the target agent $i$, computed from all other agents in the scene at time $t$ as in Figure~\ref{fig:interaction_encoder}.

The past ego motion encoder takes $e_t=(\alpha_t,\omega_t$) as input and embeds the motion history of ego-vehicle using the GRU. We use each hidden state output $h_t^e$ to compute future locations of other agents. 
The past object location encoder uses the GRU to embed the history of past motion into a feature space. The input to this module is a bounding box $x^i$ of the target agent $i$ at each past time step, and we use the embedding $\Phi(x^i_t)$ for the GRU. The output hidden state $h_t^p$ of the encoder is updated by $\hat{h_{t}^p}=\Phi(H_{t}^{xi}\boxtimes h_t^p)$, where $H_{t}^{xi}=\Phi_r(a^i_{t}) \boxtimes \psi_{t}^i \boxtimes \Phi_r(h_t^e)$ is the concatenated information. Then, $\hat{h_{t}^p}$ is used as a hidden state input to the GRU by $h_{t+1}^p=GRU(\hat{h_t^p},\Phi(x_t^i);W_{\textnormal{POL}})$, where $W_{\textnormal{POL}}$ are the weight parameters. We use its final hidden state as an initial hidden state input of the future object location decoder.

The future bounding boxes of the target agent $i$ are decoded using the GRU-based future object location decoder from time step $T_{obs}+1$ to $T_{pred}$. At each time step, we output a 10-dimensional vector where the first 5 values are the center $\mu_c=(c_u, c_v)$, variance $\sigma_c =(\sigma_{cu}, \sigma_{cv})$, and its correlation $\rho_c$ and the rest 5 values are the dimension $\mu_l=(l_u, l_v)$, variance $\sigma_l=(\sigma_{lu}, \sigma_{lv})$, and its correlation $\rho_l$. We use two bi-variate Gaussians for bounding box centers and dimensions, so that they can be independently sampled. We use the negative log-likelihood loss function as:
\begin{equation}
\begin{split}
    \mathcal{L}_O = -\frac{1}{T}\sum_{t=T_{obs}+1}^{T_{pred}} &log\ p(c| \mu_{c}^t,\sigma_{c}^t, \rho_{c}) p(l| \mu_{l}^t,\sigma_{l}^t,\rho_{l}).
\end{split}
\end{equation}


\subsection{Future Ego-motion prediction}

\begin{figure}[t!]
    \centering
    \includegraphics[width=0.47\textwidth]{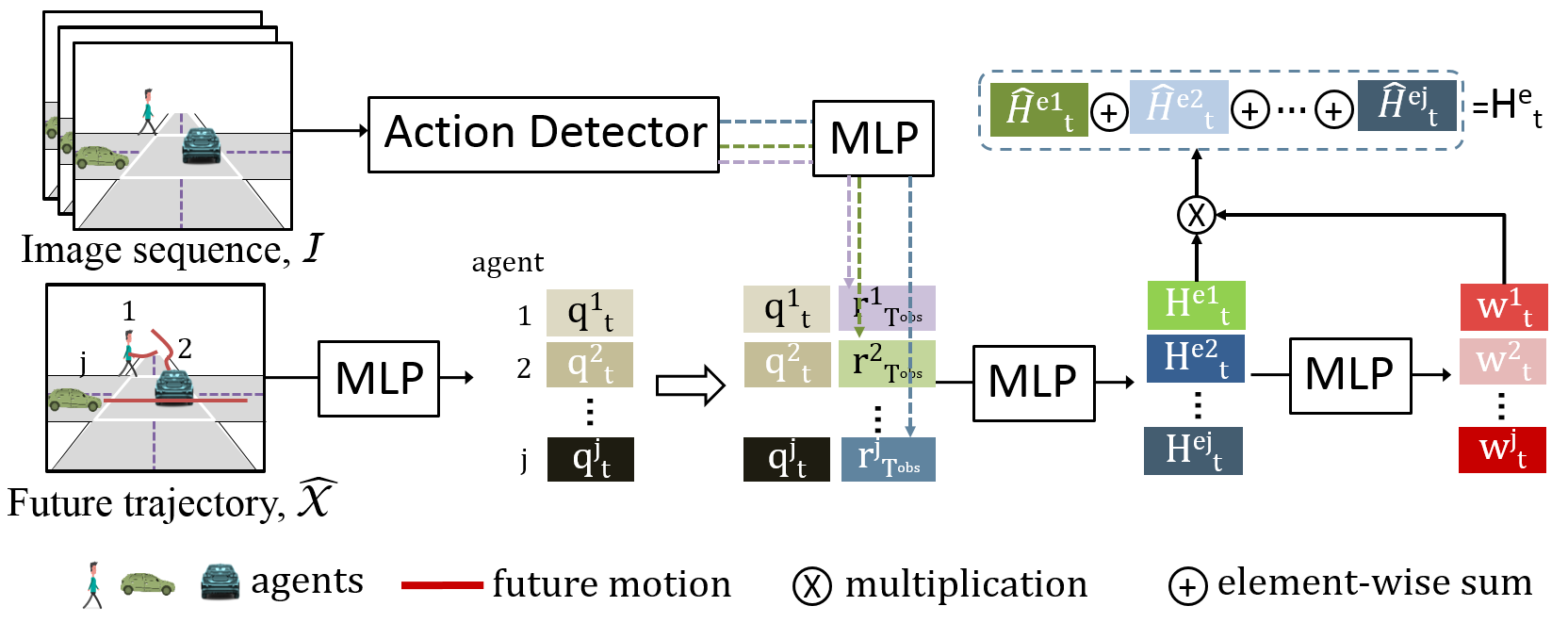}
    \caption{ Agent Importance Mechanism (AIM) module.}
    \label{fig:attention_module}
    \vspace{-0.31cm}
\end{figure}

We first embed the predicted future bounding box of all agents $\hat{X}=\{\hat{x}^1,...,\hat{x}^N\}$ through MLP at each future time step $T_{obs}+1$ to $T_{pred}$. We further condition it on the previously computed action labels in a feature space through $H_t^{ei} = \Phi(r^i_{T_{obs}}\boxtimes \Phi_r(\hat{x}_t^i))$, where $r^i_{T_{obs}}=\Phi_r(a_{T_{obs}}^i)$. By using the action labels as a prior constraint, we explicitly lead the model to understand about the contextual meaning of locations. The resulting features of each agent $i$ are weighted using the AIM module $\hat{H}_t^{ei} = w_t^i*{H}_t^{ei}$, where the weights $w_t^i = \Phi_t(H_t^{ei})$, similar to self-attention~\cite{selfattention}. Then, we sum all features $H^e_t=\sum_i \hat{H}_t^{ei}$ for each future time step. This procedure is detailed in Figure~\ref{fig:attention_module}. Note that our AIM module is simultaneously learned with the future ego-motion prediction, which results in weighting other agents more or less based on their influence/importance to the ego-vehicle. It thus provides  insight into assessment of perceived risk while predicting the future motion. We qualitatively evaluate it in Sec.~\ref{experiments}.

The last hidden state $h_{T}^e$ of the past ego motion encoder is concatenated with $H^e_t$ through $\hat{h}_{T}^e=\Phi(H^{e}_t\boxtimes h_{T}^e)$ and fed into the future ego motion decoder. The intermediate hidden state $h_t^f$ is accordingly updated by $H^e_t$ at each future time step for recurrent update of the GRU. We output the ego-future using each hidden state $h_t^f$ through $\hat{e}_{t}^i=\Phi(h_t^f)$ at each future time $T_{obs}+1$ to $T_{pred}$. For training, we use task dependent uncertainty with L2 loss for regressing both acceleration and angular velocity as shown below:
\begin{equation}
\mathcal{L}_E = \frac{\norm{\alpha_t-\hat{\alpha}_t}^2}{\sigma_1^2}+\frac{\norm{\omega_t-\hat{\omega}_t}^2}{\sigma_2^2}+log \sigma_1 \sigma_2.
\end{equation}

Note that the predicted future ego-motion is deterministic in its process. However, its multi-modality comes from sampling of the predicted future bounding boxes of other agents. In this way, we capture their influence with respect to the ego-vehicle, and AIM outputs the importance weights consistent with the agents' action and future motion. 


\begin{figure*}[t]

\begin{center}
\includegraphics[width=0.98\textwidth]{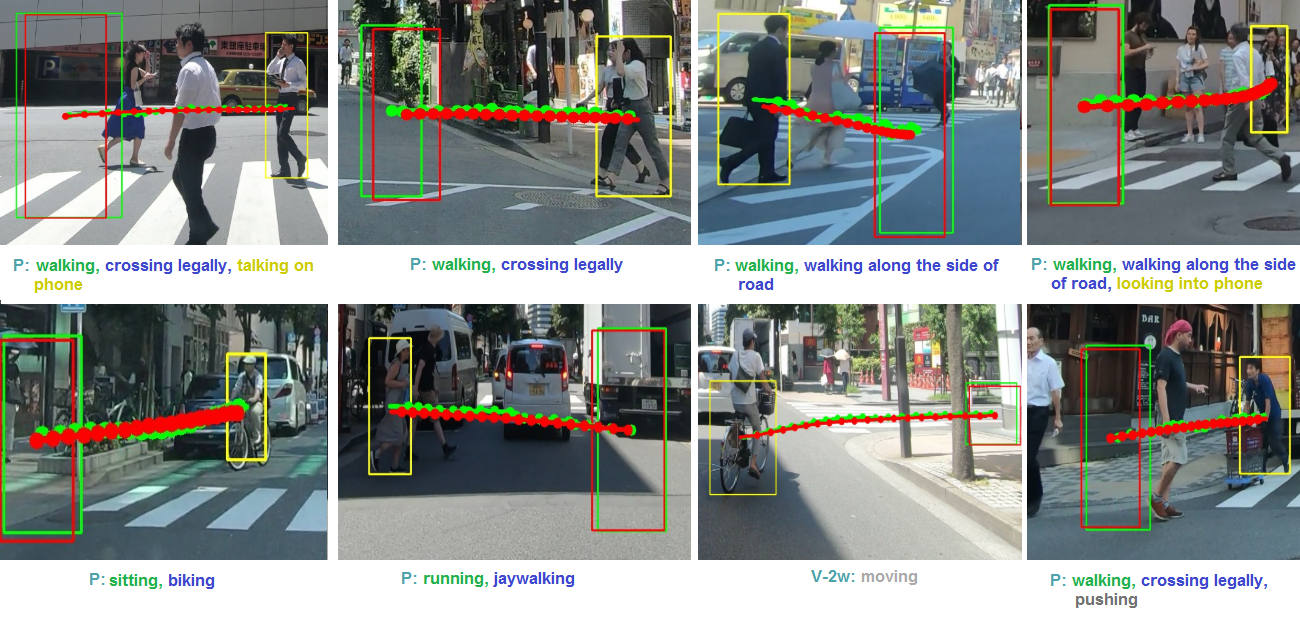}

    \end{center}
    \vspace{-0.61cm}
  \caption{Qualitative evaluation on the TITAN dataset: ground truth future trajectory \protect\inlinegraphics{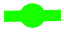}, TITAN prediction \protect\inlinegraphics{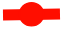}, last observation bounding box \protect\inlinegraphics{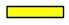}. The color of detected action labels indicates each action set described in Figure~\ref{fig:titan_dataset}. Images are cropped for better visibility.
  }\vspace{-0.11cm}
  \label{fig:ours}
\end{figure*}

\begin{figure*}
\begin{center}
\includegraphics[width=\textwidth]{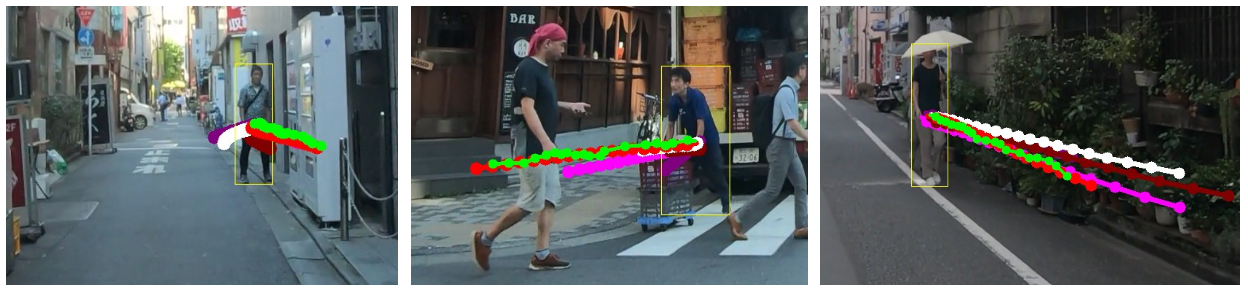}
\end{center}
\vspace{-0.51cm}
  \caption{Comparison with others: ground truth   \protect\inlinegraphics{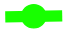}, Titan\_EP+IP+AP (ours) \protect\inlinegraphics{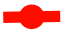}, Titan\_EP+IP (w/o action) \protect\inlinegraphics{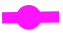}, Social-LSTM~\cite{social_lstm} \protect\inlinegraphics{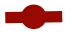}, Social-GAN~\cite{social_gan} \protect\inlinegraphics{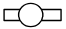}, Const-Vel~\cite{constvel} \protect\inlinegraphics{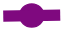}, bounding box at $T_{obs}$ \protect\inlinegraphics{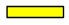}. Images are cropped for better visibility.
  }
  \label{fig:comp}
   \vspace{-0.55cm}
\end{figure*}

\section{Experiments}
\label{experiments}
In all experiments performed in this work, we predict up to 2 seconds into the future while observing 1 second of past observations as proposed in ~\cite{malla2019nemo}. We use average distance error (ADE), final distance error (FDE), and final intersection over union (FIOU) metrics for evaluation of future object localization. We include FIOU in our evaluation since ADE/FDE only capture the localization error of the final bounding box without considering its dimensions.  For action recognition, we use per frame mean average precision (mAP).  Finally, for ego-motion prediction, we use root mean square error (RMSE) as an evaluation metric.

\subsection{Action Recognition}
We evaluate two state-of-the-art 3D convolution-based architectures, I3D with InceptionV1 and 3D ResNet with ResNet50 as backbone. Both models are pre-trained on Kinetics-600 and finetuned using TITAN with the multi-task loss in Eqn.~\ref{eq:multi_task_loss}. As detailed in Sec.~\ref{sec:action_rec}, we modify the original structure using new heads that corresponds to the 8 action sets of the TITAN dataset. Their per frame mAP results are compared in Table~\ref{tbl:action_recog} for each action set. We refer to the supplementary material for the detailed comparison on individual action categories. Note that we use the I3D-based action detector for the rest of our experiments.

\begin{table}[t!]
\centering\footnotesize
\begin{tabular}{ l|l||c|c}
&Method & I3D~\cite{i3d} & \multicolumn{1}{|c}{3D ResNet~\cite{r3d}} \\
\hline\hline
&Backbone &InceptionV1&ResNet50\\
\hline
&atomic&0.9219&0.7552\\
&simple&0.5318&0.3173\\
person&complex&0.9881&0.9880\\
&communicative&0.8649&0.8648\\
&transportive&0.9080&0.9081\\\cline{2-4}
&overall&0.8429&0.7667\\
\hline
&motion&0.9918&0.7132\\
vehicle&trunk&1.00&1.00\\
&doors&1.00&1.00\\\cline{2-4}
&overall&0.9921&0.9044\\
\hline
\multicolumn{2}{c||}{overall$\uparrow$}&0.8946&0.8128\\
\hline
\end{tabular}\vspace{0.21cm}
 \caption{Action recognition results (mAP) on TITAN.}
 \vspace{-0.51cm}
  \label{tbl:action_recog}
\end{table}

\subsection{Future Object Localization}

\begin{figure*}[t!]
\begin{center}
\mbox{\subfigure{
\includegraphics[width=\textwidth]{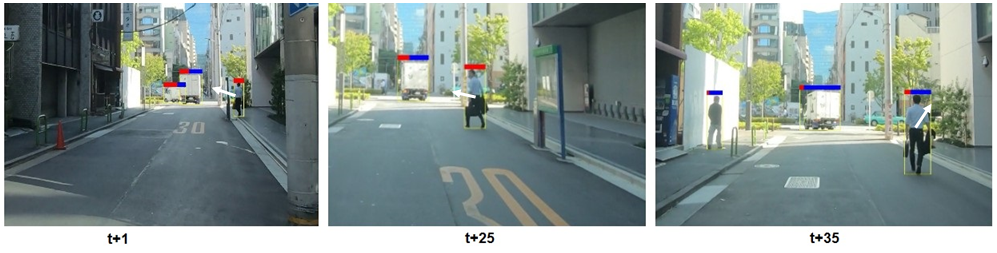}}}
\mbox{\subfigure{
\includegraphics[width=\textwidth]{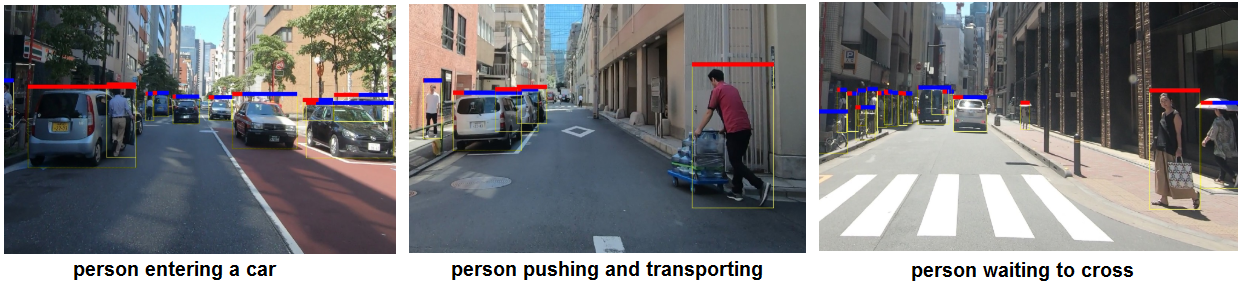}}}
    \end{center}
    \vspace{-0.31cm}
  \caption{The importance (or degree of influence) of each agent toward the ego-vehicle's future trajectory is illustrated by the proportion of red bar relative to the blue bar displayed across the top width of the agent's bounding box.  A red bar spanning across the top width represents the maximum importance derived from the AIM module, while a blue bar spanning across the top width represents minimum importance.  (top row) images from same sequence.  (bottom row) images from different sequences.
  }
  \label{fig:attn}
\end{figure*}
The results of future object localization performance is shown in Table~\ref{tbl:fol}. The constant velocity (Const-Vel~\cite{constvel}) baseline is computed using the last two observations for linearly interpolating future positions.  Since the bounding box dimensions error is not captured by ADE or FDE, we evaluate on FIOU using two baselines: 1) without scaling the box dimensions, and 2) with scaling linearly the box dimensions. Titan\_vanilla is an encoder and decoder RNN without any priors or interactions.  It shows better performance than linear models. Both Social-GAN~\cite{social_gan} and Social-LSTM~\cite{social_lstm} improve the performance in ADE and FDE compared to the simple recurrent model (Titan\_vanilla) or linear approaches. Note that we do not evaluate FIOU for Social-GAN and Social-LSTM since their original method is not designed to predict dimensions. Titan\_AP adds action priors to the past positions and performs better than Titan\_vanilla, which shows that the model better understands contextual meaning of the past motion. However, its performance is worse than Titan\_EP that includes ego-motion as priors. This is because Titan\_AP does not consider the motion behavior of other agents in egocentric view. Titan\_IP includes interaction priors as shown in Figure~\ref{fig:interaction_encoder} without concatenating actions. Interestingly, its performance is better than Titan\_AP (action priors) and Titan\_EP (ego priors) as well as Titan\_EP+AP (both ego and action priors). It validates the efficacy of our interaction encoder that aims to pass the interactions over all agents. This is also demonstrated by comparing Titan\_IP with two state-of-the-art methods. With ego priors as default input, interaction priors (Titin\_EP+IP) finally perform better than Titan\_IP.  Interactions with action information (Titan\_EP+IP+AP) significantly outperforms all other baselines, suggesting that interactions are important and can be more meaningful with the information of actions\footnote{Using ground-truth actions as a prior, we observed further improvement in overall ADE by 2 \textit{pixels} and overall FDE by 3.5 \textit{pixels}.}.

\begin{table}
\centering\footnotesize
\begin{tabular}{ l||c|c|c }
Method & ADE $\downarrow$ & FDE $\downarrow$ & FIOU $\uparrow$  \\\hline\hline
Const-Vel (w/o scaling)~\cite{constvel} & 44.39 & 102.47  & 0.1567\\
Const-Vel (w/ scaling)~\cite{constvel}  & 44.39 & 102.47& 0.1692\\ 
Social-LSTM~\cite{social_lstm} & 37.01 & 66.78 & -  \\
Social-GAN~\cite{social_gan}  & 35.41 & 69.41 & - \\
\hline
Titan\_vanilla  & 38.56 & 72.42 & 0.3233\\
Titan\_AP  & 33.54 & 55.80 & 0.3670\\
Titan\_EP  & 29.42 & 41.21 & 0.4010\\ 
Titan\_IP  & 22.53 & 32.80  & 0.5589\\ 
Titan\_EP+AP  & 26.03 & 38.78 & 0.5360\\ 
Titan\_EP+IP  & 17.79 & 27.69 & 0.5650\\ 
Titan\_EP+IP+AP (ours)  & \textbf{11.32} & \textbf{19.53} & \textbf{0.6559}\\  \hline
\end{tabular}
 \caption{Quantitative evaluation for future object localization. ADE are FDE in pixels on the original size 1920x1200.}
 \vspace{-0.51cm}
 \label{tbl:fol}
\end{table}

The qualitative results are shown in Figure~\ref{fig:ours}. The proposed method predicts natural motion for the target with respect to their detected actions (listed below each example). In Figure~\ref{fig:comp}, we compare ours with the baseline models. The performance improvement against Titan\_EP+IP further validates our use of action priors for future prediction. Additional results can be found in the supplementary material.

\begin{table}[t!]
\centering\footnotesize
\begin{tabular}{ l||c|c }
Method & acc RMSE $\downarrow$ & yaw rate RMSE $\downarrow$ \\\hline\hline
Const-Vel~\cite{constvel} & 1.745 & 0.1249 \\
Const-Acc & 1.569 & 0.1549 \\
\hline
Titan\_vanilla  & 1.201 & 0.1416  \\
Titan\_FP  & 1.236 & 0.1438 \\
Titan\_FP+AP  & 1.182 & 0.1061 \\
Titan\_AIM\_FP  & 1.134 & 0.0921 \\
Titan\_AIM (ours)  & \textbf{1.081}  & \textbf{0.0824} \\

\hline
\end{tabular}\vspace{0.21cm}
 \caption{Comparison of Future ego motion prediction. acceleration error in $m/s^2$ and yaw rate error in $rad/s$. }
 \vspace{-0.51cm}
  \label{tbl:fe}
\end{table}

\subsection{Future Ego-Motion Prediction}

The quantitative results for future ego-motion prediction are shown in Table~\ref{tbl:fe}.  
Between Const-Vel~\cite{constvel} and Const-Acc (acceleration), the Const-Vel baseline performs better in predicting angular velocity (yaw-rate) and Const-Acc performs better for predicting acceleration. Titan\_vanilla only takes the past ego-motion as input, performing better than Const-Vel and Const-Acc for acceleration prediction. Although incorporating information of other agents' future predictions (Titan\_FP) does not improve the performance over Titan\_vanilla, the addition of their action priors (Titan\_FP+AP) shows better performance for both acceleration and yaw rate prediction. By adding just future position in the AIM module (Titan\_AIM\_FP), the system can weight the importance of other agents' behavior with respect to the ego-future, resulting in decreased error rates. Finally, by incorporating future position and action in the AIM module as a prior yields the best performance, Titan\_AIM. 

To show the interpretability of which participant is more important for ego-future, we visualize the importance weights in Figure~\ref{fig:attn}.  In particular, the top row illustrates that the importance weight of the pedestrian increases as the future motion direction (in white arrow) is towards the ego-vehicle's future motion. Although the agent is closer to the ego-vehicle at a later time step, the importance decreases as the future motion changes.  This mechanism provides insight into assessment of perceived risk for other agents from the perspective of the ego-vehicle. 

\section{Conclusion}
We presented a model that can reason about the future trajectory of scene agents from egocentric views obtained from a mobile platform. Our hypothesis was that action priors provide meaningful interactions and also important cues for making future trajectory predictions. To validate this hypothesis, we developed a model that incorporates prior positions, actions, and context to simultaneously forecast future trajectory of agents and future ego-motion.   For evaluation, we created a novel dataset with over 700 video clips containing labels of a diverse set of actions in urban traffic scenes from a moving vehicle. Many of those actions implicitly capture the agent's intentions. Comparative experiments against baselines and state-of-art prediction algorithms showed significant performance improvement when incorporating action and interaction priors.  Importantly, our framework introduces an Agent Importance Mechanism (AIM) module to identify agents that are influential in predicting the future ego-motion, providing insight into assessment of perceived risk in navigation.  For future work, we plan to incorporate additional scene context to capture participant interactions with the scene or infrastructure.

\textbf{Acknowledgement} 
We thank Akira Kanehara for supporting our data collection and Yuji Yasui, Rei Sakai, and Isht Dwivedi for insightful discussions.

{\small
\bibliographystyle{ieee}
\bibliography{egbib}
}


\newpage
\phantom{a}

\begin{figure*}[t]
\includegraphics[width=0.99\textwidth]{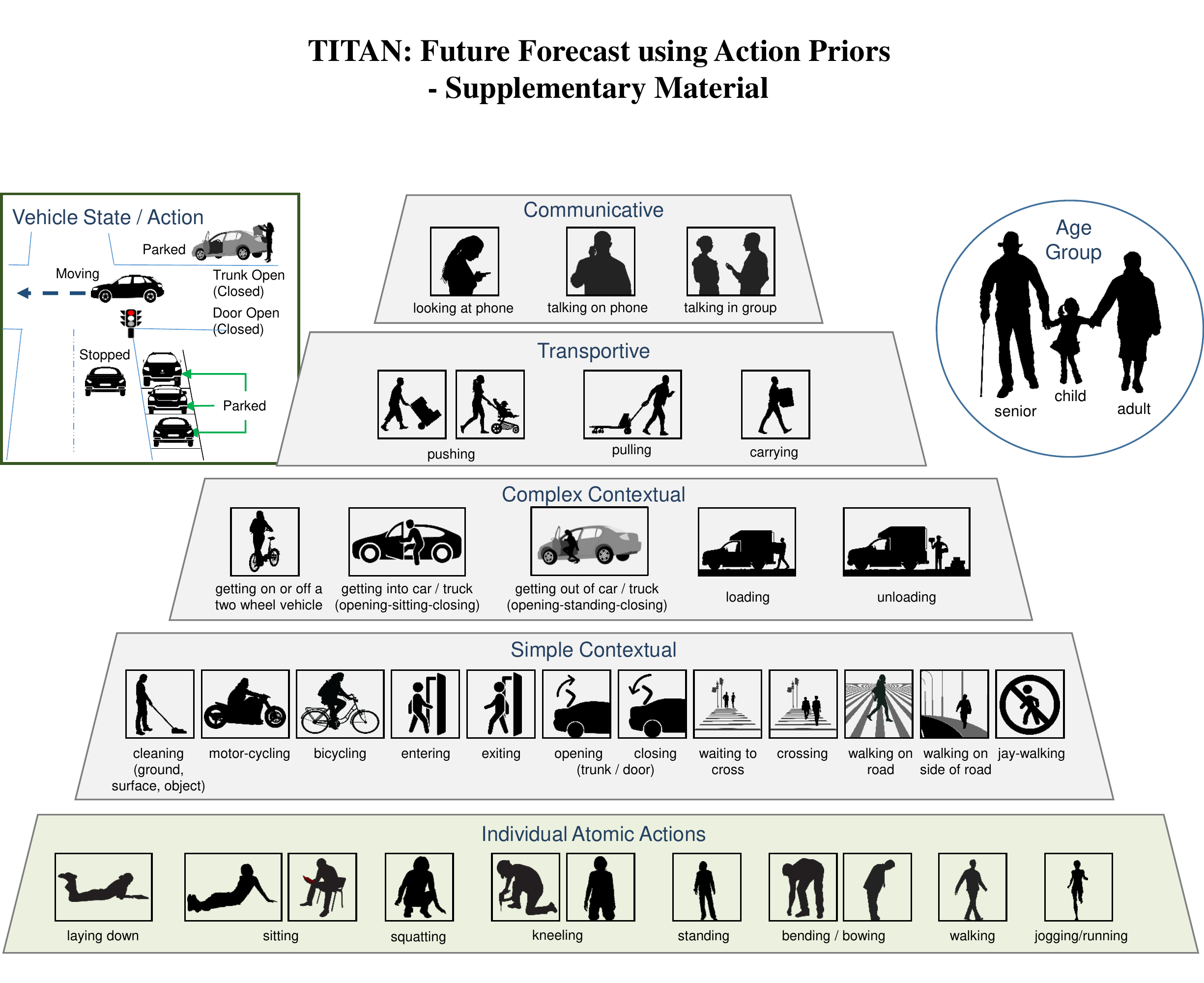}\vspace{0.3cm}
\caption{Our TITAN dataset contains 50 labels including vehicle states and actions, pedestrian age groups, and targeted pedestrian action attributes that are organized hierarchically corresponding to atomic, simple/complex-contextual, transportive, and communicative actions.}
\label{fig:main}
\end{figure*}
\newpage

\begin{table*}
\centering
\begin{tabular}{ c|c|c|c|l}
Category&Set&\# Classes & \# Instances & Description\\
\hline\hline
 &Atomic&9&392511&atomic whole body actions/postures that describe primi-\\
&&&&tive action poses (\textit{e.g.}, sitting, standing, walking, etc.)\\\cline{2-5}
&Simple contextual&13&328337&single atomic actions that include scene context (\textit{e.g.},\\
&&&&jaywalking, waiting to cross)\\\cline{2-5}
Human&Complex contextual&7&5084&a sequence of atomic actions with increased complexity\\
Action&&&&and/or higher contextual understanding\\\cline{2-5}
&Transportive&4&35160&manually transporting an object by carrying, pulling, or\\
&&&& pushing.\\\cline{2-5}
&Communicative&4&57030&communicative actions (e.g. talking on the phone, look-\\
&&&&ing at phone, or talking in groups.)\\\hline
&Motion status&3&249080&motion status of 2-wheeled and 4-wheeled vehicles\\
&&&&(parked / moving / stationary)\\\cline{2-5}
Vehicle& Trunk status &2&146839&trunk for 4-wheeled vehicles\\
State&&& &(open / closed)\\\cline{2-5}
&Door status &2&146839&door status for 4-wheeled vehicles\\
&&& &(open / closed)\\\hline
&Age group &3&395769&subjective categorization of pedestrians into age groups \\
Other&&&&(child / adult / senior)\\\cline{2-5}
Labels&Object type&3&645384&participant types categorized into \\
 &&&&pedestrian / 2-wheeled / 4- wheeled vehicles\\
\hline
\end{tabular}
\caption{Details of the TITAN dataset. We report the number of labels, instances, and descriptions for each action set.} \label{tbl:titan_dataset}
\end{table*}
\section{Details of the TITAN Dataset}
Figure~\ref{fig:main} illustrates the labels of the TITAN dataset, which are typically observed from on-board vehicles in driving scenes. We define 50 labels including vehicle states and actions, pedestrian age groups, and targeted pedestrian action attributes that are hierarchically organized from primitive atomic actions to complicated contextual activities. Table~\ref{tbl:titan_dataset} further details the number of labels, instances, and descriptions for each action set in the TITAN dataset. For pedestrians, we categorize human actions into 5 sub-categories based on their complexities and compositions. Moreover, we annotate vehicle states with 3 sub-categories of motion, and trunk / door status. Note that the trunk and door status is only annotated for 4-wheeled vehicles.  Vehicles with 3-wheels without trunk but with doors are annotated as 4-wheeled and trunk open. Also, 3-wheeled vehicles with no trunk and doors are annotated as 2-wheeled vehicles.  The list of classes for human actions is shown in Table~\ref{tbl:per_class_results}. The annotators were instructed to only localize pedestrians and vehicles with a minimum bounding box size of $70\times10$ $pixels$ and $50\times10$ $pixels$ in the image, respectively.

Several example scenarios of TITAN are depicted in Figure~\ref{fig:titan_dataset}. In each scenario, four frames are displayed with a bounding box around a road agent. We also provide their actions below each frame. Note that only one agent per frame is selected for the purpose of visualization. The same color code is used for each action label, which can be found in Figure 2 of the main manuscript.

\section{Additional Evaluation}
In this section, we provide additional evaluation results of the proposed approach.
\subsection{Per-Class Quantitative Results}
In Table~\ref{tbl:per_class_results}, we present per-class quantitative results of the proposed approach, which are evaluated using the test set of TITAN. Note that the number of instances for some actions (\textit{e.g.}, \textit{kneeling}, \textit{jumping}, etc.) are zero, although they are present in the training and validation set. This is because we randomly split 700 clips of TITAN into training, validation, and test set. We will regularly update TITAN to add more clips with such actions.

We observe that the error rate for some classes are either much lower or higher than other classes. For example, scenarios depicting \textit{getting into a 4 wheel vehicle}, \textit{getting out of a 4 wheel vehicle}, and \textit{getting on a 2 wheel vehicle} show very small FDE as compared to others. Also, scenarios depicting \textit{entering a building} has a larger ADE and FDE than other scenarios.  The reason for this can be explained by considering interactions of agents. When a person is \textit{getting into a vehicle}, the proposed interaction encoder builds a pair-wise interaction between the person (subject that the action generates) and the vehicle (object that the subject is related to). It further validates the efficacy of our interaction encoding capability. In contrast, no interactive object is given to the agent for \textit{entering a building} class since we assume agents are either pedestrians or vehicles. As mentioned in the main manuscript, we plan to incorporate additional scene context such as topology or semantic information.

\subsection{Efficacy of Multi-Task Loss}

The comparative results of the I3D action recognition module with and without the multi-task (MT) loss is shown in Table~\ref{tbl:mt_loss}. The performance improvement for atomic and simple contextual actions for pedestrians and motion status for vehicles with the MT loss validates its efficacy of modeling aleatoric homoscedastic uncertainty of different tasks.

\begin{figure*}[t!]
\begin{center}
\includegraphics[width=0.99\textwidth]{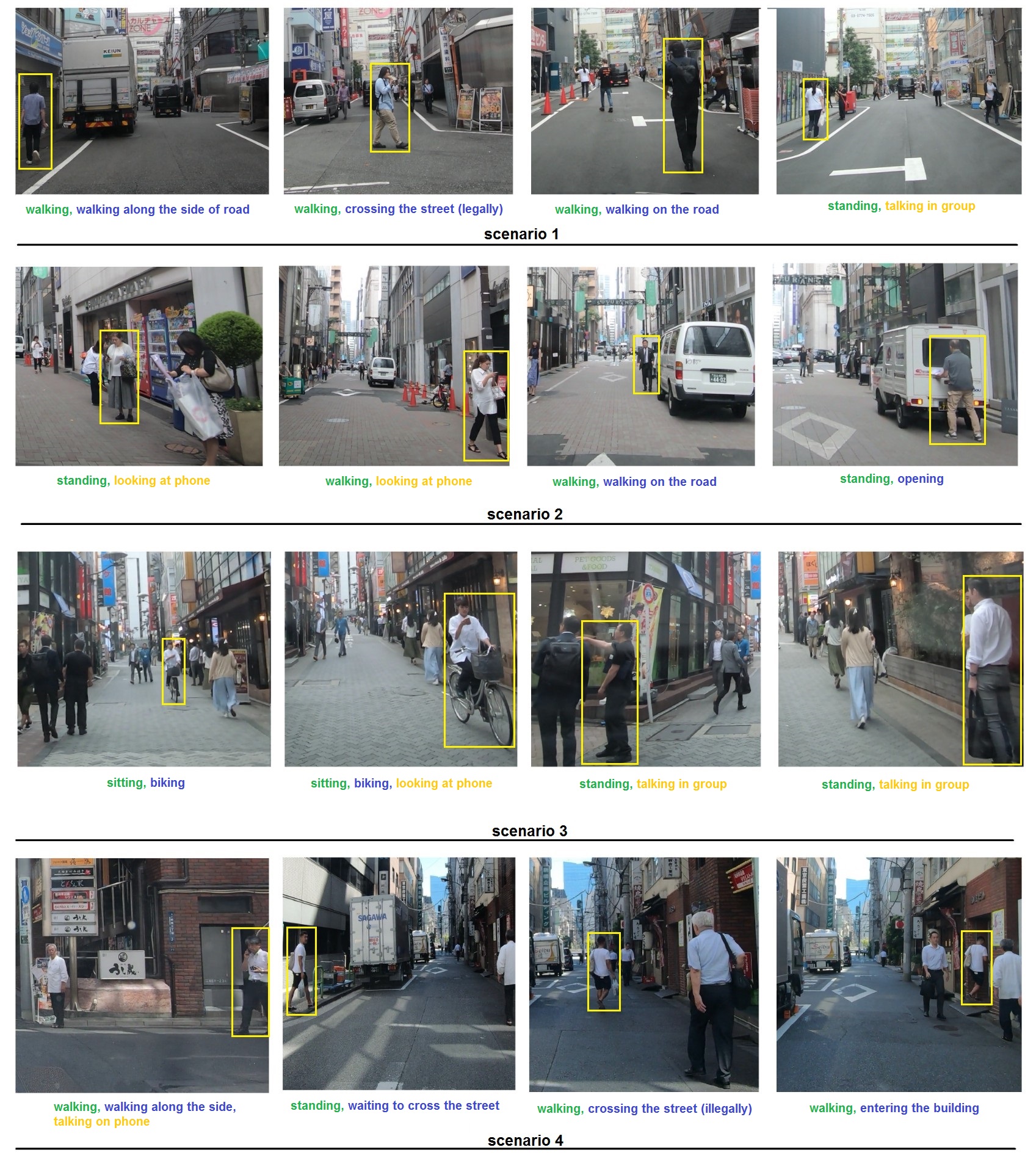}
\vspace{0.3cm}
\caption{Example sequences from the TITAN dataset. Some notable actions are highlighted with different color codes following the hierarchy in the main manuscript (Color codes: Green - atomic, Blue - simple contextual, and Yellow - communicative). Images are cropped from their original size for better visibility.}
\label{fig:titan_dataset}
\end{center}
\end{figure*}

\begin{table*}[t!]
\begin{center}
\begin{tabular}{ c|c|c|c|c|c}
Action Set& Class&ADE$\downarrow$ & FDE$\downarrow$ & FIOU$\uparrow$ & \#Instances\\
\hline\hline
&standing&10.56&18.63&0.6128&1476\\
&running&12.39&19.95&0.6179&89\\
&bending&12.76&20.85&0.6560&156\\
&kneeling&0.00&0.00&0.00&0\\
Atomic Action&walking&13.31&23.15&0.5712&6354\\
&sitting&11.10&20.74&0.6282&337\\
&squatting&11.90&18.82&0.5598&4\\
&jumping&0.00&0.00&0.00&0\\
&laying down&0.00&0.00&0.00&0\\
&none of the above&9.69&16.43&0.7408&7237\\
\hline
&crossing at pedestrian crossing& 13.22 &21.59 &0.5976 & 881\\
&jaywalking & 13.10 &21.91 &  0.6148 & 340\\
&waiting to cross street& 11.49 &21.75 & 0.5783 & 65\\
&motorcycling&20.00 &31.81 & 0.5494 & 4\\
&biking&13.22 &21.13 & 0.6283 & 287\\
&walking along the side of the road& 11.33 &24.50 & 0.5516 & 2668\\
Simple-Contextual&walking on the road& 13.41 &22.30 & 0.5794 & 2486\\
&cleaning (ground, surface, object)& 11.67&22.58 & 0.6502 & 19\\
&closing&9.84 & 20.50& 0.4947 & 14\\
&opening&12.99 &29.89 &0.1995 & 13\\
&exiting a building&13.56 & 28.09&0.5264 & 61\\
&entering a building&28.06 & 53.02 & 0.2259 & 6\\
&none of the above& 9.85 & 16.76& 0.7201 & 8809 \\
\hline
&unloading& 11.07 & 18.45& 0.7082 &37 \\
&loading&11.59 & 18.54& 0.6652 & 40\\
&getting in 4 wheel vehicle&8.39 &10.80 &0.5682 &10\\
Complex-Contextual&getting out of 4 wheel vehicle& 9.63 & 9.58& 0.7972 &3\\
&getting on 2 wheel vehicle& 7.73 &11.16 & 0.7619&10\\
&getting off 2 wheel vehicle& 0 & 0& 0 &0\\
&none of the above& 11.32 & 19.54 & 0.6557 & 15553 \\
\hline
&looking at phone& 12.12 & 21.48  & 0.6435 &  392  \\
Communicative&talking on phone& 11.69  & 19.39 & 0.6056 &  268 \\
&talking in group& 11.70 & 20.82  & 0.6025 &   461 \\
&none of the above& 11.28 & 19.43  & 0.6588  & 14532  \\
\hline
&pushing&12.57  & 23.07  & 0.6148  &   232  \\
Transportive&carrying with both hands& 11.39  & 20.23 & 0.6477  &  445 \\
&pulling& 12.01 & 21.29 & 0.5198 & 76 \\
&none of the above& 11.29  & 19.44 & 0.6574  &  14900 \\
\hline\hline
&stopped&8.96 & 23.08 & 0.6148 &232\\
Motion-Status&moving& 9.18&20.23 &0.6477  &  445\\
&parked&9.93 &21.29 & 0.5199 &  76\\
&none of the above& 12.72&19.44 &0.6574  &  14900\\
\hline
\end{tabular}
 \vspace{0.3cm}
 \caption{Per-class evaluation results using the test set of 100 clips.}
 \vspace{1.51cm}
  \label{tbl:per_class_results}
\end{center}
\end{table*}

\begin{figure*}[t!]
    \centering
    \includegraphics[width=0.99\textwidth]{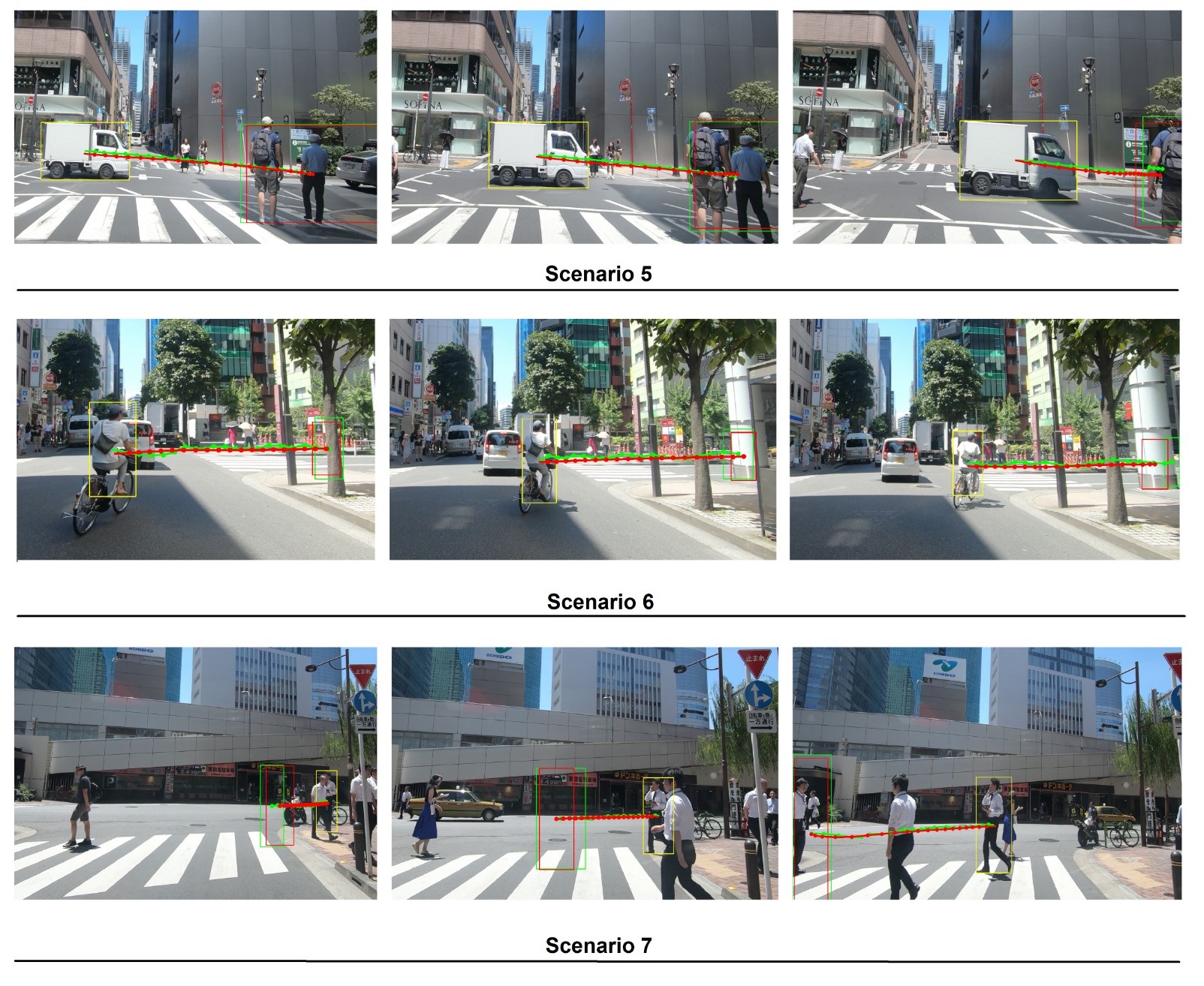}
    \vspace{0.4cm}
    \caption{Qualitative results of TITAN from different sequences.   Trajectories in Red denote predictions, trajectories in green color denote ground truth, and a yellow bounding box denotes the last observations. (Images cropped for better visibility)}
    
    \label{fig:ours}
\end{figure*}
\begin{figure*}[h!]
    \centering
    \includegraphics[width=0.99\textwidth]{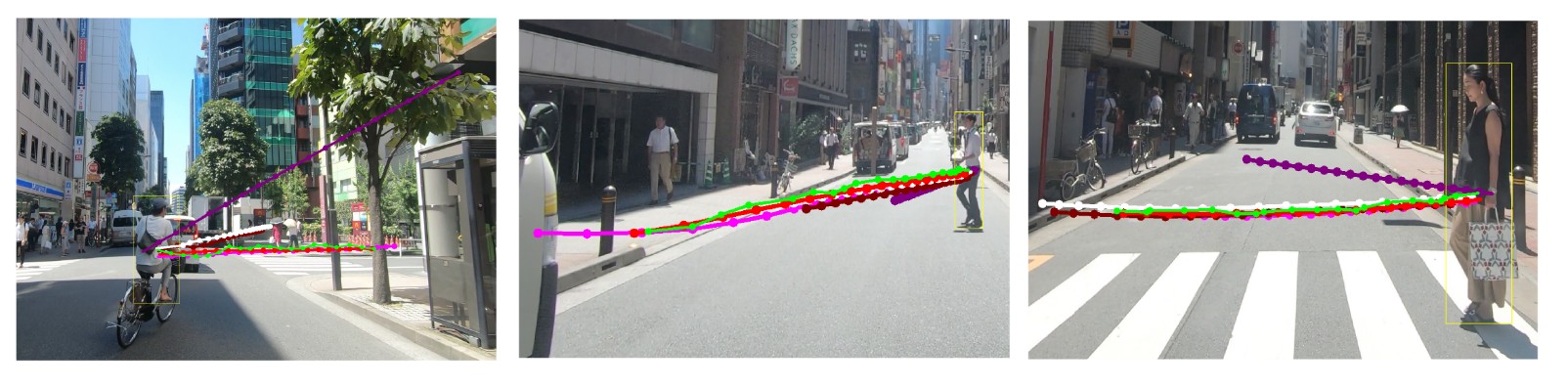}
    \vspace{0.4cm}
    \caption{Comparison with others: ground truth   \protect\inlinegraphics{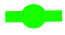}, Titan\_EP+IP+AP (ours) \protect\inlinegraphics{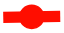}, Titan\_EP+IP (w/o action) \protect\inlinegraphics{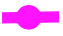}, Social-LSTM~\cite{social_lstm} \protect\inlinegraphics{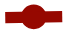}, Social-GAN~\cite{social_gan} \protect\inlinegraphics{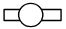}, Const-Vel~\cite{constvel} \protect\inlinegraphics{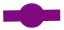}, bounding box at $T_{obs}$ \protect\inlinegraphics{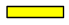}. Images are cropped for better visibility.}
    \label{fig:comparison}
\end{figure*}

\begin{figure*}[!h]
    \centering
    \includegraphics[width=0.95\textwidth]{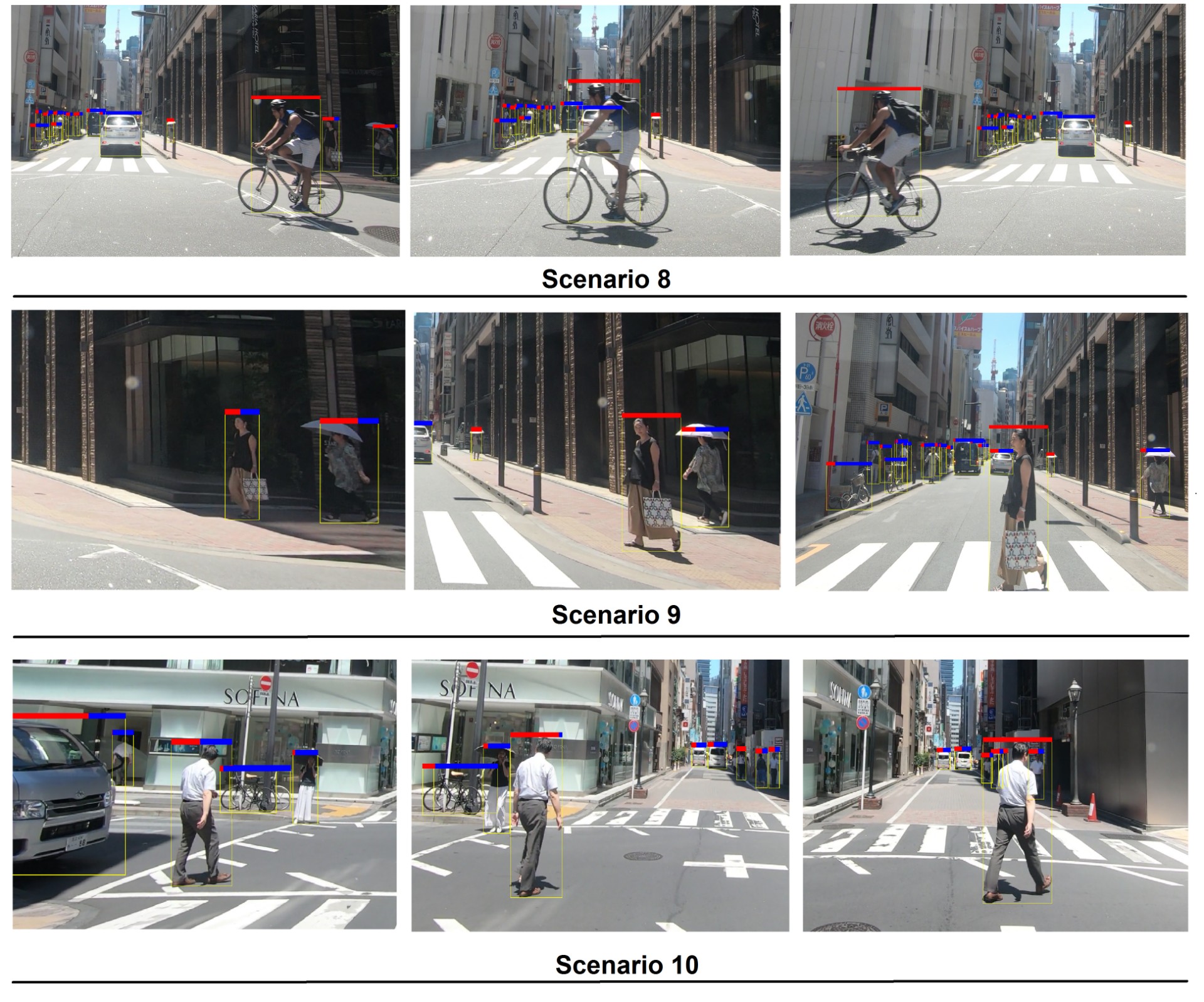}
    \caption{Qualitative results of Importance from different sequences, RED color is high importance, blue is low importance and yellow bounding box is the last observation. (Images cropped for better visibility)}
    \label{fig:attn}
\end{figure*}

\begin{table}[ht!]
\centering
\small
\begin{tabular}{ l|l||c|c}
&Method & w/ MT loss$\uparrow$ & w/o MT loss$\uparrow$\\
\hline
\hline
\parbox[t]{2mm}{\multirow{5}{*}{\rotatebox[origin=c]{90}{person}}}&atomic&0.9219&0.7552\\
&simple&0.5318&0.3173\\
&complex&0.9881&0.9880\\
&communicative&0.8649&0.8647\\
&transportive&0.9080&0.9080\\\cline{2-4}
&overall&0.8429&0.7667\\
\hline
\parbox[t]{2mm}{\multirow{3}{*}{\rotatebox[origin=c]{90}{vehicle}}}&motion&0.9918&0.7130\\
&trunk&1.00&1.00\\
&doors&1.00&1.00\\\cline{2-4}
&overall&0.9921&0.9043\\
\hline
\multicolumn{2}{c||}{overall$\uparrow$}&0.8946&0.8127\\
\hline
\end{tabular}
 \vspace{0.4cm}
 \caption{Action recognition results (mAP) on TITAN.}
  \label{tbl:mt_loss}
\end{table}

\subsection{Additional Qualitative Results}
Figure~\ref{fig:ours} and \ref{fig:comparison} show the prediction results of the proposed approach for future object localization. Titan\_EP+IP+AP consistently shows better performance against the baseline model and the state-of-the-art methods. We also observed that t

In Figure~\ref{fig:attn}, the proposed Agent Importance Module (AIM) is evaluated on additional sequences. The ego-vehicle decelerates due to the crossing agent, and our system considers this agent as having a higher influence (or importance)than other agents.   Agents with high importance are depicted with a red over-bar.
Particularly in scenario 10, when the person walks along the road in the longitudinal direction, its importance is relatively low. However, the importance immediately increases when the motion changes to the lateral direction.

\section{Implementation}
TITAN framework is trained on a Tesla V100 GPU using PyTorch Framework. We separately trained action recognition, future object localization, and future ego-motion prediction modules. During training, we used ground-truth data as input to each module. However, during testing, the output results of one module are directly used for later tasks.

\subsection{Future Object Localization}
During training, we used a learning rate of 0.0001 with RMSProp optimizer and trained for 80 epochs using a batch size of 16. We used hidden state dimension of 512 for both encoder and decoder. A size of 512 is used for the embedding size of action, interaction, ego-motion and bounding box. The input box dimension is 4, action dimension is 8, and ego-motion dimension is 2. The original image size width is 1920 $pixels$ and height is 1200 $pixels$ and accordingly cropped using the bounding box dimension. It is further resized to $224\times224$ for the I3D-based action detector. The bounding box inputs and outputs are normalized between 0 to 1 using image dimensions. 


\begin{table}[h!]
\centering\scriptsize
\begin{tabular}{ c|l|c|c}
 &Layer & Kernal shape & Output shape\\
\hline\hline
0&ego$\_$box$\_$embed.Linear$\_$0 & [4, 512] & [1, 10, 512]\\
1&ego$\_$box$\_$embed.ReLU$\_$1 & - &  [1, 10, 512]\\
2&ego$\_$action$\_$embed.Linear$\_$0 & [8, 512] & [1, 512]\\
3&ego$\_$action$\_$embed.ReLU$\_$1 & - & [1, 512]\\
4&ego$\_$motion$\_$embed.Linear$\_$0 & [2, 512]& [1, 10, 512]\\
5&ego$\_$motion$\_$embed.ReLU 1 & - & [1, 10, 512]  \\ 
6&box$\_$encoder.GRUCell$\_$enc & - & [1, 512]   \\
7&motion$\_$encoder.GRUCell$\_$enc & - & [1, 512]  \\ 
8&int$\_$encoder.embed.Linear$\_$0&[24, 512]&[1, 512] \\  
9& int$\_$encoder.embed.ReLU$\_$1 & - & [1, 512]  \\ 
10& int$\_$encoder.encode.GRUCell$\_$enc&  - &		[1, 512]  \\ 
11& concat$\_$to$\_$hidden.Linear$\_$0  &	 [2048, 512] &	[1, 512] \\
12& concat$\_$to$\_$hidden.ReLU$\_$1&- &	[1, 512]\\   \hline
13&  \multicolumn{3}{|c}{(8 to 12, repeat based on number of pairwise interactions)}\\\hline
14& \multicolumn{3}{|c}{(0 to 12, unroll 10 times)}\\\hline
15& pred.GRUCell$\_$dec & - & [1, 512]   \\
16& pred.hidden$\_$to$\_$input.Linear$\_$0&   [512, 512] &	[1, 512] \\
17& pred.hidden$\_$to$\_$input.ReLU$\_$1  & -  &	 [1, 512] \\  
18& pred.hidden$\_$to$\_$output.Linear$\_$0 &  [512, 10] &	[1, 10]  \\ 
19& pred.hidden$\_$to$\_$output.Sigmoid$\_$1 & -&	 [1, 10]\\   \hline
20& \multicolumn{3}{|c}{(15 to 19, unroll 20 times)}\\
\hline
\end{tabular}
 \caption{Future Object Localization model summary with an example batch size of 1}
  \label{tbl:fol_model}
\end{table}
The model summary for Future Object Localization is shown in Table~\ref{tbl:fol_model}. We embed the bounding box (through 0 and 1), action (2-3), ego-motion (4-5) at each time step, and pairwise interaction encoding (8-12). We concatenate the embedded features through (11-12), which are given from the hidden states of the bounding box encoder GRU (6), the hidden states of the ego encoder GRU (7), encoded interaction (10) and action embedding (3). We encode all information for 10 observation time steps from (14). We decode the future locations using decoder GRU for 20 future time steps (20).

\subsection{Action Recognition}
We used Kinetics-600 pre-trained weights for both I3D and 3D-ResNet. For I3D, we use layers until Mixed$\_$5c layer of the original structure. We used learning rate of 0.0001 and a batch size of 8. We trained it for 100 epochs. The input size is $3\times10\times224\times224$, where 10 is the number of time steps, 3 is the number of RGB channels. If the agent is occluded and reappears at any time step, we used the last observed crop of image for that the agent. During training we backpropagate the gradients for pedestrians and vehicles with the loss function as shown below:
\begin{equation}
    \mathcal{L}_{total}=\mathds{1}_p {\mathcal{L}_a}^{i=1:5}+(1-\mathds{1}_p){\mathcal{L}_a}^{i=6:8},
    \label{eq:multi_task_loss_sep}
\end{equation}
where $\mathds{1}_p$ is an indicator function that equals 1 if the agent is a pedestrian and 0 if the agent is a vehicle. We refer to the main manuscript for $\mathcal{L}_a$. 
\begin{table}[h!]
\centering\scriptsize
\begin{tabular}{ c|l|c|c}
&Layer & Kernal shape & Output shape\\
\hline\hline
1&i3d.Conv3d$\_$1a$\_$7x7.conv3d&[3, 64, 7, 7, 7]&[1, 64, 5, 112, 112]\\\hline
..& \multicolumn{3}{|c}{.....}\\\hline
126&i3d.Mixed$\_$5c.b3b.BatchNorm3d&[128]&[1, 128, 2, 7, 7]  \\\hline
127&action.hid$\_$to$\_$pred1.Linear$\_$0&[100352, 10]&[1, 10] \\
128&action.hid$\_$to$\_$pred1.Softmax$\_$1& -&[1, 10]\\ 
129&action.hid$\_$to$\_$pred2.Linear$\_$0&[100352, 13]&[1, 13]\\
130&action.hid$\_$to$\_$pred2.Softmax$\_$1&- &[1, 13] \\
131&action.hid$\_$to$\_$pred3.Linear$\_$0&[100352, 7]&[1, 7] \\
132&action.hid$\_$to$\_$pred3.Softmax$\_$1& -&[1, 7] \\
133&action.hid$\_$to$\_$pred4.Linear$\_$0&[100352, 4]&[1, 4] \\
134&action.hid$\_$to$\_$pred4.Softmax$\_$1&-&[1, 4] \\
135&action.hid$\_$to$\_$pred5.Linear$\_$0&[100352, 4]&[1, 4]\\
136&action.hid$\_$to$\_$pred5.Softmax$\_$1& - &[1, 4]\\
137&action.hid$\_$to$\_$pred6.Linear$\_$0 &[100352, 4]&[1, 4]\\
138&action.hid$\_$to$\_$pred6.Softmax$\_$1 & -&[1, 4] \\
139&action.hid$\_$to$\_$pred7.Linear$\_$0&[100352, 3] &[1, 3]\\ 
140&action.hid$\_$to$\_$pred7.Softmax$\_$1 & - & [1, 3]\\
141&action.hid$\_$to$\_$pred8.Linear$\_$0 & [100352, 3] & [1, 3]\\
142&action.hid$\_$to$\_$pred8.Softmax$\_$1 & - & [1, 3]\\
\hline
\end{tabular}
 \caption{I3D action recognition model summary with an example batch size of 1}
  \label{tbl:action_model}
\end{table}
The model summary for action recognition is shown in Table~\ref{tbl:action_model}. Note that, from mixed$\_$5c layer [b0, b1b, b2b, b3b] are concatenated to give a shape of [1,1024,2,7,7] which is flattened to give a tensor of shape [1,100352] before feeding it to each MLP head for individual action sets.
\subsection{Future Ego-Motion Prediction}
\begin{table}[h]
\centering\scriptsize
\begin{tabular}{ c|l|c|c}
&Layer & Kernal shape & Output shape\\
\hline\hline
0&ego$\_$embed.Linear$\_$0&[2, 128]&[1, 10, 128] \\  
1&ego$\_$embed.ReLU$\_$1& - &[1, 10, 128]  \\ 
2&ego$\_$encoder.GRUCell$\_$enc & - & [1, 128] \\  \hline
3&\multicolumn{3}{|c}{(0 to 3, unroll 10 times)}\\\hline
4&pred.box$\_$embed.Linear$\_$0&[4, 128]&[1, 1, m, 128]\\
5&pred.box$\_$embed.ReLU$\_$1& - & [1, 1, m, 128]\\  
6&pred.action$\_$embed.Linear$\_$0 & [8, 128] & [1, 1, m, 128] \\
7&pred.action$\_$embed.ReLU$\_$1 & - & [1, 1, m, 128] \\
8&pred.concat$\_$to$\_$hid2.Linear$\_$0 & [256, 128] & [1, 1, m, 128]\\   
9&pred.AIM$\_$layer.Linear$\_$0 & [128, 1]& [1, 1, m, 1] \\ 
10&pred.AIM$\_$layer.Tanh$\_$1& - & [1, 1, m, 1] \\  
11&pred.concat.concat$\_$0 &-& [1, 256] \\  
12&pred.concat$\_$to$\_$hid.Linear$\_$0 &[256, 128]& [1, 128] \\  
13&pred.GRUCell$\_$dec & - & [1, 128] \\  
14&pred.hid$\_$to$\_$pred$\_$input.Linear$\_$0 & [128, 128]& [1, 128] \\
15&pred.hid$\_$to$\_$pred$\_$input.ReLU$\_$1 &  - &	[1, 128] \\  
16&pred.Linear$\_$hid$\_$to$\_$pred & [128, 2]& [1, 2]  \\\hline
17&\multicolumn{3}{|c}{(4 to 15, unroll 20 times)}\\
\hline
\end{tabular}
 \caption{Future ego motion prediction model summary with an example batch size of 1, m is the number of agents at that future time step}
  \label{tbl:fe_model}
\end{table}

We use batch size of 64, learning rate of 0.0001 and trained for 100 epoch with RMSProp optimizer. We use the hidden state dimension of 128 for both encoder and decoder. We use the embedding size of 128. The prediction is done for 20 time steps in future. The input and output dimensions are 2 at each time step.

The model summary of the future ego-motion prediction is shown in Table~\ref{tbl:fe_model}.
We embed the ego motion at each time step (0-1) and use GRU encoder (2) for 10 observation time steps (3). The encoded information is used for the decoder. The embedded future bounding box (4-5) and embedded current action (6-7) are concatenated (8). The agent importance module (AIM) is used to weight the agents at each time step (9-10). We concatenate (11) the AIM output with the past hidden state and embed it (12). The embedded feature is used as an input hidden state. The current hidden state (13) is passed to the next time-step (14-15) using GRU. The output is decoded (16) from the hidden state at each time step (17). As a result, we get for 20 future predictions.


\end{document}